\documentclass[letterpaper, 10 pt, conference]{ieeeconf}  % Comment this line out if you need a4paper

\IEEEoverridecommandlockouts                              % This command is only needed if 
\title{\LARGE \bf
Cascade Attribute Learning Network
}

\author{Zhuo Xu*$^{1}$, Haonan Chang*$^{2}$, and Masayoshi Tomizuka$^{1}, \emph{Fellow, IEEE}$% <-this % stops a space
\thanks{* Both authors contributed equally to this work}% <-this % stops a space
\thanks{$^{1}$Zhuo Xu and Masayoshi Tomizuka are with the Dept. of Mechanical Engineering, University of California, Berkeley CA 94720, USA \newline
        {\tt\small \{zhuoxu,tomizuka\}@berkeley.edu}}%
\thanks{$^{2}$Haonan Chang is with the Dept. of Mechanical Engineering, Tsinghua University, Beijing 100084, China \newline
        {\tt\small changhn14@mails.tsinghua.edu.cn}}%
}
\usepackage{algorithm,algpseudocode}
\usepackage{graphicx}
\usepackage{amsfonts}
\usepackage{float}
\usepackage{comment}
\begin{document}
\maketitle
\thispagestyle{empty}
\pagestyle{empty}

%%%%%%%%%%%%%%%%%%%%%%%%%%%%%%%%%%%%%%%%%%%%%%%%%%%%%%%%%%%%%%%%%%%%%%%%%%%%%%%%
\begin{abstract}

We propose the cascade attribute learning network (CALNet), which can learn attributes in a control task separately and assemble them together. Our contribution is twofold: first we propose attribute learning in reinforcement learning (RL). Attributes used to be modeled using constraint functions or terms in the objective function, making it hard to transfer. Attribute learning, on the other hand, models these task properties as modules in the policy network. We also propose using novel cascading compensative networks in the CALNet to learn and assemble attributes. Using the CALNet, one can zero shoot an unseen task by separately learning all its attributes, and assembling the attribute modules. We have validated the capacity of our model on a wide variety of control problems with attributes in time, position, velocity and acceleration phases.

\end{abstract}

%%%%%%%%%%%%%%%%%%%%%%%%%%%%%%%%%%%%%%%%%%%%%%%%%%%%%%%%%%%%%%%%%%%%%%%%%%%%%%%%
\section{INTRODUCTION}

Reinforcement learning (RL) \cite{sutton_book} has been successful in solving many control problems rooted in fixed Markov Decision Processes (MDPs) environments \cite{nn_policy}\cite{gae}\cite{visuomotor}. However, the extremely close interactions between the RL algorithms and the MDPs leads to the difficulty to reuse the knowledge learned from one task in new tasks. This difficulty further impedes RL policies from being adept in solving high dimensional complicated tasks. For example, it is easy to train an autonomous vehicle to travel from an origin position to a target position. However, if one takes into consideration a bunch of vehicle and pedestrian obstacles, the difficulty of the problem could grow overwhelming for shallow policy models. In order to avoid all the obstacles, one would have to train a deep policy network with very sparse reward input. Therefore, the training process usually requires an unbearably large amount of computation. What makes it worse is that such policies are hardly reusable in other scenarios, even if the new task is very similar to the previous one. Suppose a speed limit requirement is added to the autonomous driving task, although the input of the policy network is already tediously high dimensional, given there is no entrance for the speed limit information to enter the policy network, it is not possible that the pretrained policy accomplishes the new task, no matter how the policy network is tuned. Therefore, RL frameworks with fixed policy models can hardly address such high dimensional and complicated tasks in environments of great variance.

We propose to address this problem from a new perspective: modularizing complicated and high dimensional problems using a series of attributes. The attributes refer especially to global characteristics or requirements that take effect throughout the task. An example of attribute learning is shown in Fig. 1. Concretely, to solve the complicated driving problem, one first decompose the requirements of the task into a target reaching attribute, an obstacle avoidance attribute and a speed limit attribute, then train the modular network for each of the attributes, and finally assemble the attribute networks together to produce the overall policy. Modularizing a task using a series of attributes has three main intriguing advantages:

\begin{figure}[t]
\centering
\includegraphics[scale=0.39]{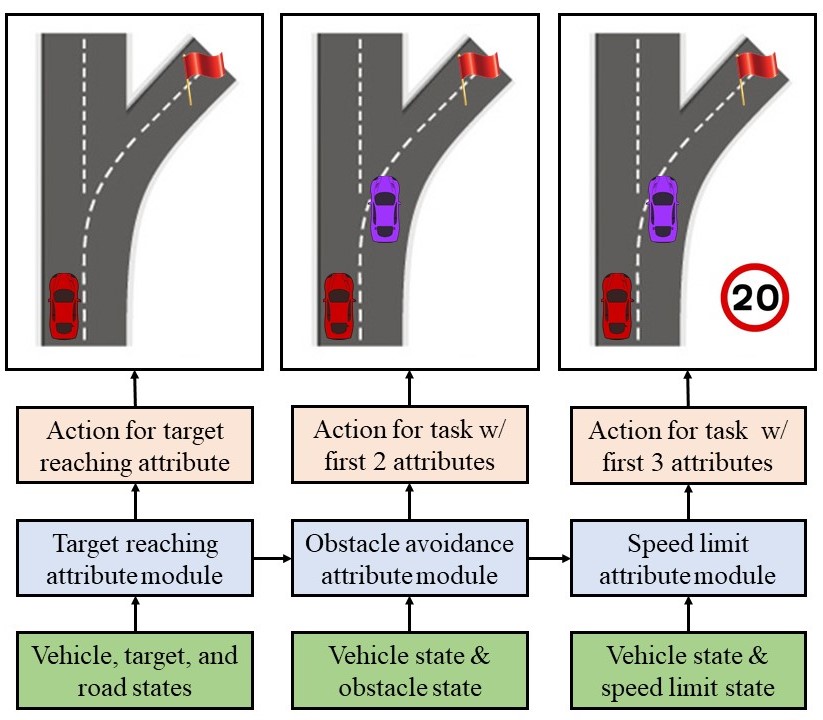}
\caption{Autonomous driving as an example of modularizing a complicated task into multiple attributes using the cascade attribute learning network (CALNet).}
\label{figurelabel}
\end{figure}

\begin{enumerate}
\item Decomposing a high dimensional complicated task into low dimensional attributes makes the training process much easier and faster.
\item Trained attribute modules can be reused in new tasks, making it possible to build up versatile policies that can adjust to changes in tasks by assembling attribute modules.
\item In attribute learning, specific state information is provided only to its corresponding attribute modules. This decoupling formulation makes it possible to dynamically manage state space in high dimensional environments.
\end{enumerate}

In order to modularize the attributes, we propose a simple but efficient RL framework called the cascade attribute learning network (CALNet). The brief idea of the CALNet is shown in Fig. 1. In CALNet, the attribute modules are connected in cascade series. Each attribute module receives both the output of its preceding module and its corresponding states, and returns the action that satisfies all the attributes ahead of it. The details of the CALNet architecture and the training methods are described in Section III. Using the CALNet, one can zero shoot an unseen task by separately learning all the attributes in the task and assemble the attribute modules in series together. The reminder of this paper is organized as follows: the related works and the background of RL are introduced in Section II. In Section III, the architecture of the CALNet and the implementation details are described. In Section IV, we show simulation results to validate the proposed model using a variety of robots and attributes and give discussions on the experiments. The conclusions are given in Section V.

\section{Related Work and Background}

\subsection{Related Work}

There have been lots of attempts to create versatile intelligence that can not only solve complicated tasks, but adjust to changes in the tasks as well. Transfer learning \cite{reinforcement_transfer}\cite{transfer} is a key tool that makes the use of previously learned knowledges for the better or faster learning of new knowledges. Rusu et al \cite{progressive1}\cite{progressive2} designed a multi-column (network) framework, referred to as progressive network, in which newly added columns are laterally connected to previously learned columns for knowledge transfer. Drafty et al \cite{transferable_policy} and Braylan et al \cite{reuse_module} also designed interesting network architectures for knowledge transfer in MAV control and video game playing. For the combinations of transfer learning and imitation learning, Ammar et al \cite{unsupervised_transfer} uses unsupervised learning to map states for transfer, assuming the existence of distance function between different state spaces. Gupta et al \cite{invariant_feature} learns an invariant feature between different dimensional states and use demonstrations to increase the density of the rewards. Our work differs from those works mainly in that we put emphasis on the modularization of attributes, which are concrete and meaningful modules that can be conveniently assembled into various combinations.

There are other methods seeking to learn a globally general policy: Meta learning \cite{meta} attempts to build self-adaptive learners that improve their bias through accumulating experience. One shot imitation learning \cite{one_shot}, for example, is a meta learning framework which is trained using a number of different tasks so that new skills could be learned from a single expert demonstration. Curriculum learning (CL)\cite{curriculum} trains a model on a sequence of cognate tasks that get more and more challenging gradually, so as to solve hard tasks that could not be learned from scratching. Florensa et al \cite{reverse_curriculum} applied reverse curriculum generation (RCL) in RL. In the early stage of the training process, the RCL initializes the agent state to be very close to the target state, making the policy very easy to train. They then gradually increase the random level of the initial state as the RL model performs better and better. Our policy training strategy is inspired by the idea of CL and achieved satisfying robustness for the policies. There are also researches in training modular neural networks, \cite{modular_robot_task} investigates the combinations of multiple robots and tasks, while \cite{modular_subtask} investigates the combinations of multiple sequential subtasks. Our work, different from those works, looks into modularization in a different dimension. We investigate the modularization of attributes, the characteristics or requirements that take effect throughout the whole task.

\subsection{Deep Reinforcement Learning Background}

The objective of RL is to maximize the expected sum of the discounted rewards $R_t = \mathbb{E} \sum_{k=0}^\infty \gamma^{k} \cdot r_{t+k}$ in an agent-environment-interacting MDP. The agent observes state $s_t$ at time $t$, and selects an action $a_t$ according to its policy $\pi_\theta$ parameterized by $\theta$. The environment receives $s_t$ and $a_t$, and returns the next state $s_{t+1}$ and the reward in this step $r_t$. The $\gamma$ in the objective function is a discounting coefficient. The main approaches for reinforcement learning include deep Q-learning (DQN) \cite{dqn}, asynchronous advantage actor critic (A3C) \cite{a3c}, trust region policy optimization (TRPO) \cite{trpo}, and proximal policy optimization (PPO) \cite{ppo}. Approaches used in continuous control are mostly policy gradient methods, i.e. A3C, TRPO, and PPO. Vanilla policy gradient method updates the parameters $\theta$ by ascending the log probability of action $a_t$ with higher advantage $\hat{A_t}$. The surrogate objective function is
$$
L(\theta) = \hat{\mathbb{E}}_t \left[ \log \pi_{\theta}(a_t \mid s_t) \cdot \hat{A_t} \right] \eqno{(1)}
$$
Although A3C uses the unbiased estimator of policy gradient, large updates can prevent the policy from converging. TRPO introduces a constraint to restrict the updated policy from being too far in Kullback-Leibler (KL) distance \cite{kl} from the old policy. Usually, TRPO solves an unconstrained optimization with a penalty punishing the KL distance between $\pi_{\theta}$ and $\pi_{\theta_{old}}$, specifically,
$$
L(\theta) = \hat{\mathbb{E}}_t \left[ \frac{ \pi_{\theta}(a_t \mid s_t)}{\pi_{\theta_{old}}(a_t \mid s_t)} \cdot \hat{A_t} -\beta \cdot \textrm{KL} \left( \pi_{\theta} , \pi_{\theta_{old}} \right) \right] \eqno{(2)}
$$
However, the choice of the penalty coefficient $\beta$ has been a problem \cite{ppo}. Therefore, PPO modifies TRPO by using a simple clip function parameterized by $\epsilon$ to limit the policy update. Specifically,
$$
L(\theta) = \hat{\mathbb{E}}_t \left[ \min \left( \frac{ \pi_{\theta}}{\pi_{\theta_{old}}}, \textrm{clip} \left(\frac{ \pi_{\theta}}{\pi_{\theta_{old}}}, 1+\epsilon, 1-\epsilon \right) \right)\hat{A_t} \right] \eqno{(3)}
$$
This simple objective turns out to perform well while enjoying better sample complexity, thus we are using PPO as the default RL algorithm in our policy training. We are also inspired by \cite{dppo} to build a distributed framework with multiple threads to speed up the training process.

The advantage function $\hat{A_t}$ describes how better a policy is compared to a baseline. Traditionally the difference between the estimated Q value and value functions is applied as the advantage \cite{a3c}. Recently Schulman et al \cite{gae} proposed using generalized advantage estimation (GAE) to leverage the bias and variance of the advantage estimator.

\section{The CALNet}

\subsection{Problem Formulation}
We consider an agent performing a complicated task with multiple attributes. Since the agent is fixed, its action space is a fixed space, which we call $A$. We decompose the task into a series of attributes, denoted $\{ 0,1,2,\ldots \}$. We refer to the $0^{th}$ attribute as the base attribute, which usually corresponds to the most fundamental goal of the task, such as the target reaching attribute in the autonomous driving task.  We define the state space of each attribute to be the minimum state space that fully characterizes the attribute, denoted $S=\{S_0,S_1,S_2,S_3\ldots \}$. For example, let the base attribute be the target reaching attribute, and the $1^{st}$ attribute be the obstacle avoidance attribute. Then $S_0$ consists of the states of the agent and the target, while $S_1$ consists of the states of the agent and the obstacle, and yet does not include the states of the target.

Each attribute has an unique reward function as well, denoted $R=\{R_0,R_1,R_2,R_3\ldots\}$. Each $R_i$ is a function mapping a state action pair to a real number reward, i.e. $R_i: S_i \times A \rightarrow \mathbb{R}$. Similarly, there is a specific transition probability distribution for each attribute, denoted: $P=\{P_0, P_1, P_2, P_3 \ldots \}$. And for each attribute, its transition function takes in the state action pairs and outputs the states for the next timestep, that is, $P_i : S_i \times A \rightarrow S_i$.

A key characteristic in our problem formulation is that the state spaces for different attributes can be different. This formulation enables the attribute learning network to dynamically manage the state space of the task. Specifically, the states of the $i^{th}$ attribute, $s_i$, is fed to the module of the $i^{th}$ attribute in the network.

\subsection{Network architecture}
The architecture of the CALNet is shown in Fig. 2. and Fig. 3. Both the training phase (Fig. 2.) and the testing phase (Fig. 3.) of the CALNet are implemented in cascade orders. In the training phase, first a RL policy $\pi_0$ is trained to accomplish the goal of the base attribute. The base attribute network takes in $s_0\in S_0$ and outputs $a_0 \in A$, the reward and transition functions of the MDP are given by $R_0$ and $P_0$. This process is a default RL training process. 

Then the $1^{st}$ attribute module is trained in series of the base attribute module. The $1^{st}$ attribute module consists of a compensate network and a weighted sum operator. The compensate network is fed with state $s_1 \in S_1$, and action $a_0$ chosen by $\pi_0$. The output of the compensate network is the compensate action $a^{c}_1$, which is used to compensate $a_0$ to produce the overall action $a_1$. The reward for the MDP is given by $R_0+R_1$ so that the requirements for both attributes are satisfied. The new transition function may not be directly calculated using $P_0$ and $P_1$, but it can be easily obtained from the environment. Since the parameters of the base attribute network are pretrained, the cascading attribute network would extract the features of the attribute by exploring the new MDP under the guidance of the base policy.

It is noted that in the weighted sum operator, the weight of the compensative action $a^{c}_1$ is initiated to be small and increased over the training time. That is, at the early stage of the training process, mainly $a_0$ takes effect, while $a^{c}_1$ gradually gets to influence the overall $a_1$ as the training goes on. For other attributes, the training method is the same with that of the $1^{st}$ attribute.

In the testing phase, the designated attribute modules are connected in series following the base attribute, as shown in Fig. 3. In the CALNet, the $i^{th}$ attribute module takes in $s_i$ and $a_{i-1}$, and outputs $a_i$ that satisfies all the attributes before the $i^{th}$ module. The final output $a_j$ is the overall output that satisfies all the attributes in the attribute array.

\begin{figure}[t]
\centering
\includegraphics[scale=0.35]{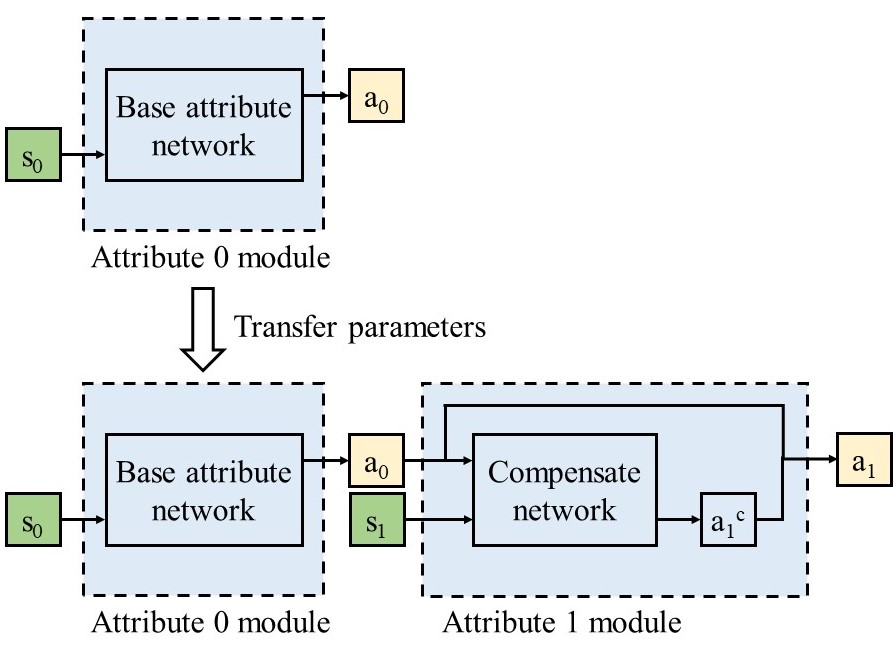}
\label{figurelabel}
\caption{The training procedure of an attribute module in the CALNet: first train the base attribute module, then train the added module based on the pretrained base module}
\end{figure}

\begin{figure}[t]
\centering
\includegraphics[scale=0.38]{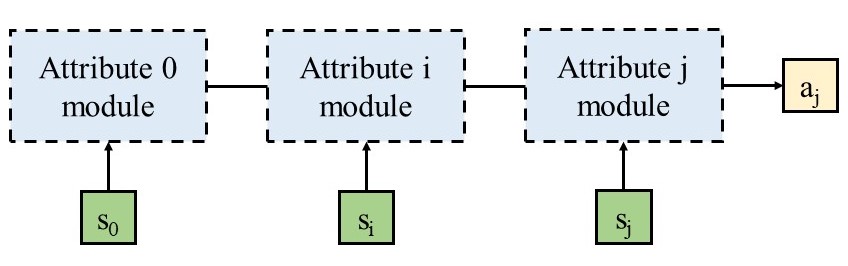}
\label{figurelabel}
\caption{Usage of the CALNet in assembling attributes onto the base attribute, the output action of the last attribute module satisfies the requirements of all the attributes}
\end{figure}

\subsection{Training Method}

To guarantee the capacity of the CALNet, the policies need to meet two requirements:

\begin{enumerate}
\item The attribute policies should be robust over the state space, rather than being effective only at the states that are close to the optimal trajectory. This requirement guarantees the attribute policies to be instructive when more compensate actions are added on the top of them.
\item The compensate action for a certain attribute should be close to zero if the agent is in a state where this attribute is not active. This property increases the capability of multi-attribute structures.
\end{enumerate}

For the sake of the robustness of the attribute policies, we apply CL to learn a general policy that can accomplish the task starting from any initial state. The CL algorithm first trains a policy with fixed initial state. As the training goes on, the random level of the initial state is smoothly increased, until the initial state is randomly sampled from the whole state space. The random level is increased only if the policy is capable enough for the current random level.

For example, consider the task of moving a ball to reach a target point in a 2 dimensional space. In each episode, the initial position of the ball is randomly sampled in a circular area. The random level in this case is the radius of the circle. In the early training stage, the radius is set to be very small, and the initial position is almost fixed. As the policy gains more and more generality, the reward in each episode increases. Once the reward reaches a threshold, the random level is increased, and the initial position of the ball is sampled from a larger area. The terminal random level corresponds to the circumstance where the circular sampling area fully covers the working zone. If the policy performs well under the terminal random level, the policy is considered successfully trained. The pseudocode for this process is shown in Algorithm 1. 

\begin{algorithm}[t]
\caption{Curriculum Learning}
\label{RCL}
\begin{algorithmic}[1]
\State $RandomLevel = $ Initial Random Level
\State $\lambda = 1 + $ Random Level Increase Rate
\State $N = $ Batch Number
\State $LongTermR = Queue()$
\While{$RandomLevel < \textrm{Terminal Random Level}$}
	\State Update the policy using PPO
    \State $Rewards \gets RunEpisode(N)$ 
    \State $LongTermR.append(Rewards)$
    \If{$Average(LongTermR) > Threshold$}
       \State $RandomLevel = RandomLevel \times \lambda$
       \State $Clear(LongTermR)$
    \EndIf
\EndWhile
\end{algorithmic}
\end{algorithm}

To guarantee the second requirement, an extra loss term that punishes the magnitude of the compensative action, $l^{c}_i \propto -\|a^{c}_i\|^2$, is added to the reward function so as to reduce $\|a^{c}_i\|$ when attribute $i$ is not active.

\section{Experiments}
Our experiments aim to validate the capability and advantage of the CALNet. In this Section, first we introduce the experiment setup, we then show the capability of the CALNet to modularize and assemble attributes in multi-attribute tasks. In the last part of this section, we compare the CALNet with the baseline RL algorithm, and show that the CALNet can adjust to complicated tasks more easily. 

\subsection{Setup}
The experiments are powered by the MuJoCo physics simulator \cite{mujoco}. The policy functions in our experiments are Gaussian distributed obtained using fully connected neural networks, built using TensorFlow. The baseline RL algorithm we use is the PPO \cite{ppo} method with GAE \cite{gae} as the advantage estimator. 

We design three robots as agents in our experiments. They are a robot arm in 2 dimensional space, a moving ball in 2 dimensional space, and a robot arm in 3 dimensional space. For all three robots we have enabled both position control and force control modes.

For each agent we have designed 5 attributes:

\begin{figure}[t]
\centering
\includegraphics[scale=0.35]{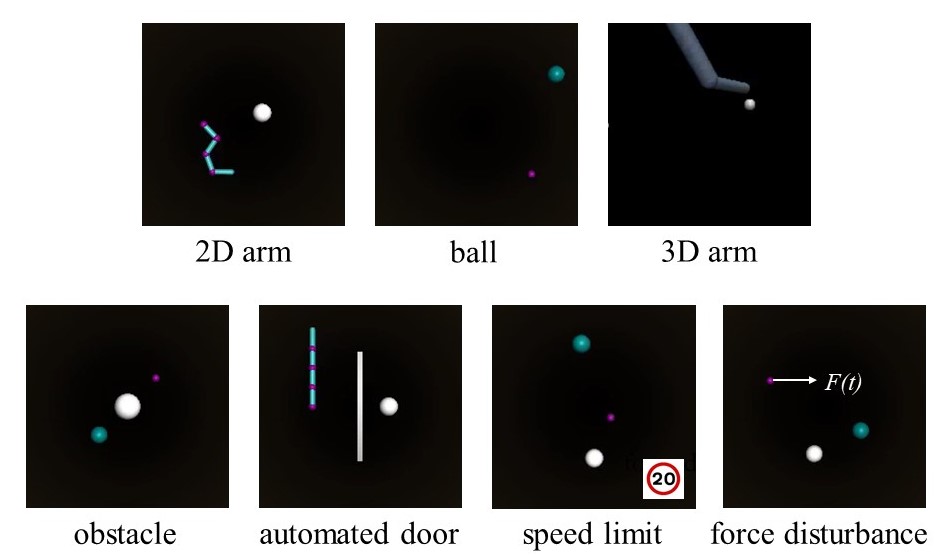}
\label{figurelabel}
\caption{The top images show the robot agents performing the base task of target reaching. The bottom images show the four attributes to be modularize.}
\end{figure}

\subsubsection{reaching (base attribute)}
The reaching task is a natural selection for the base attribute. For the ball agent, the goal is to collide the target object. For the robot arm agents, the goal is to touch the target object. 

\subsubsection{obstacle (position phase)}
The obstacle attribute is to add an rigid obstacle ball in the space. Negative rewards are given if the robot collides the obstacle. Therefore, in baseline RL training, the agent can be dissuaded from exploring the right direction.

\subsubsection{automated door (time phase)}
The automated door attribute is purely time controlled. The door blocking the target is opened only at some certain time. This attribute is harder than an obstacle, since it punishes the agent even if it goes to the right direction at a wrong time.

\begin{figure*}[t]
\centering
\includegraphics[scale=0.5]{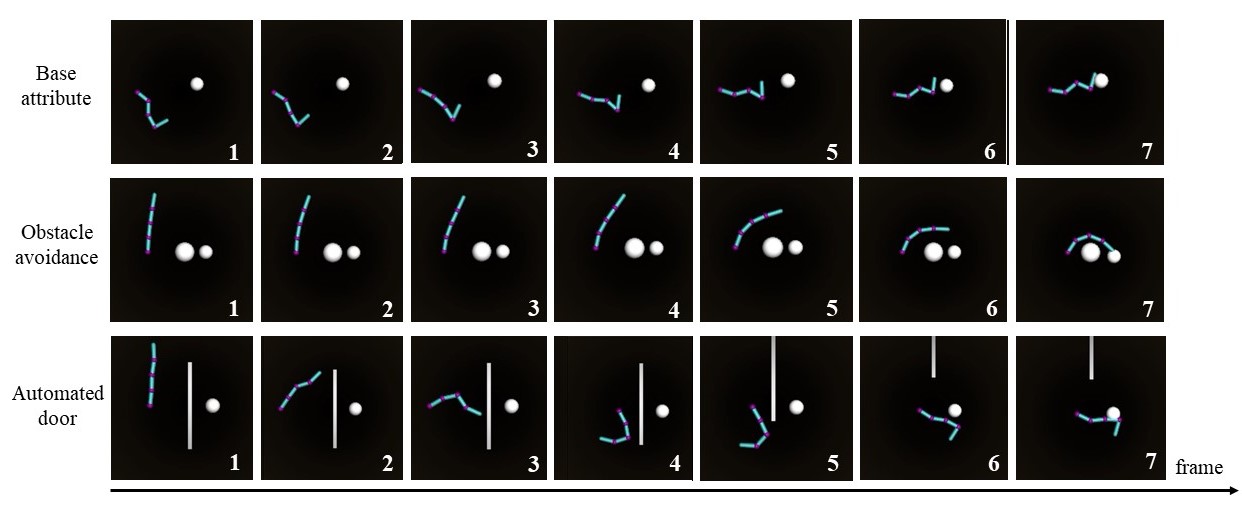}
\label{figurelabel}
\caption{The 2D arm robot performing three different tasks. The first task is the base target reaching task, the second task is the base attribute plus the obstacle avoidance attribute, the third task is the base attribute plus the time controlled door attribute. Note that there is no heuristic, and thus the reward is extremely sparse in the space.}
\end{figure*}
\begin{figure*}[t]
\centering
\includegraphics[scale=0.5]{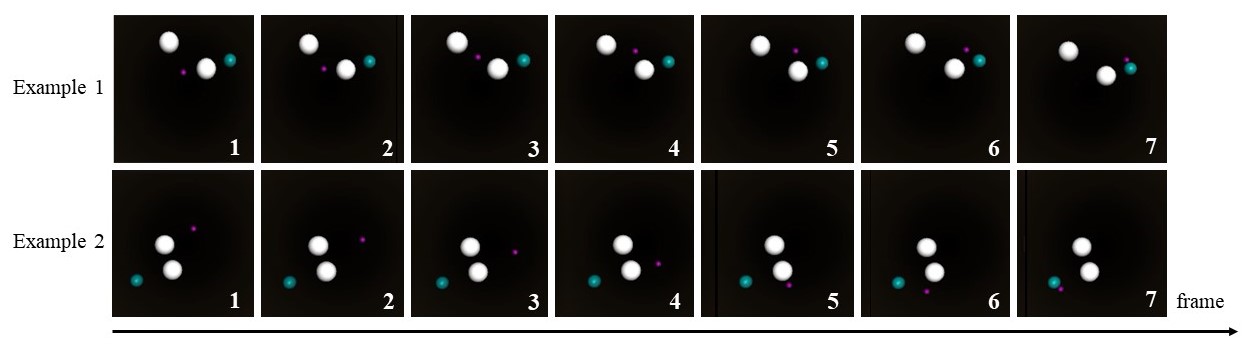}
\label{figurelabel}
\caption{Moving ball robot (the pink ball) reaches the target (the green ball) while avoiding two obstacles simultaneously. Note that this task is never seen by the CALNet. We simply train the obstacle module, and assemble two modules together and zero shoot the new task.}
\end{figure*}

\subsubsection{speed limit (velocity phase)}
The speed limit attribute adds a time-variant speed limit on the agent. The agent gets punished if it surpasses the speed limit. But if the robot's speed is too slow, it may not be able to finish the task in one episode.

\subsubsection{force disturbance (acceleration phase)}
The force disturbance attribute adds a time-variant force disturbance to the agent (or each joint for the arm).

\subsection{CALNet Performance}
The first set of experiments test the capability of the CALNet to learn attributes and assemble learned attributes. We first train the base attribute module using the baseline RL algorithm with CL, and then use the cascading modules to modularize the different attributes based on the pretrained base module. The results show that all the attributes can be successfully added to the base attribute using the CALNet. Fig. 5. shows some of the examples of the agent performing different attributes combinations.

We also test the transferability of the cascading modules and the capability of the CALNet of modeling tasks with multiple attributes. Concretely, we first train two attribute modules in parallel based on the pretrained base module. Then we connect the two attribute modules in series following the base attribute module. The CALNet structure is the same as the one shown in Fig. 3. The policy derived using the assembled network can zero shoot most of the tasks satisfying requirements of both attributes. 

Fig. 6. shows two examples of the CALNet zero shooting a task where the moving ball reaches the target while avoiding two obstacles simultaneously. It is emphasized that this task is never trained before. We achieve zero shooting simply by connecting two pretrained obstacle attribute modules in series following the base module. Undeniably as the attributes grow more complicated and the number of attributes gets larger, it would require a certain amount of finetune. However, the advantage of modularizing and assembling attributes is remarkable, since the finetuning process is much easier and faster compared to training a new policy from scratch (as discussed in Section IV-C).

\subsection{Comparison with Baseline RL Methods}
We compare the capability of the CALNet and the baseline RL by comparing their training processes on a same task. We consider the MDP in which the ball agent gets to the target while avoiding an obstacle. The CALNet is trained with CL. For baseline RL trained with CL, in many cases it is to too hard for the agent to reach the target. Therefore, we have also implemented RCL, which let the initial state be very close to the target in the early stage of the training phase. Using RCL, the RL could gain positive reward very fast. The challenge would be whether the RL algorithm could maintain high reward level as the random level increases.

For CALNet, the base attribute has been trained, and we train the obstacle avoidance attribute module based on the base module. For the baseline RL, the task is trained from scratching. The focus of the comparison is placed on the responding reward and random level in CL versus the training iterations.

The reward and the random level curves are shown in Fig. 7, with the horizontal axis representing the training iterations. It is shown that the baseline RL using CL barely learns anything. This is because the reward is too sparse and the agent is consistently receiving punish from the obstacle, and fells into some local minimum. For the baseline RL using RCL, in the early stage, the average discounted reward in an episode is high as expected. But as the random level rises, the performance of the baseline RL with RCL drops. Therefore, the random level increases slowly as the training goes on.

The CALNet, on the other hand, is able to overcome the misleading punishments from the obstacle, thanks to the guidance of the instructive base attribute policy. As a result, the random level of the CALNet rises rapidly, and the CALNet achieves terminal random level more than 10 times faster than the baseline. These results indicate that the attribute module learns substantial knowledge of the attribute as the CL based training goes on.

\begin{figure}[t]
\centering
\includegraphics[scale=0.7]{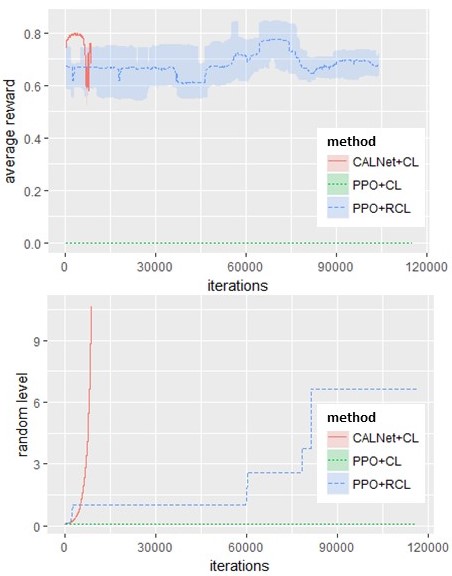}
\label{comparision}
\caption{Comparison between the performance of the CALNet and the baseline RL (PPO) in the training phase.}
\end{figure}

\begin{comment}
\begin{table}[h]
\caption{An Example of a Table}
\label{table_example}
\begin{center}
\begin{tabular}{|c||c|}
\hline
One & Two\\
\hline
Three & Four\\
\hline
\end{tabular}
\end{center}
\end{table}
\end{comment}

\section{Conclusions}

In this paper, we propose the attribute learning and present the advantages of using this novel method to modularize complicated tasks. The RL framework we propose, the CALNet, uses cascading attribute modules to model the characteristics of the attributes. The attribute modules are trained with the guidance of the pretrained base attribute module. We validated the effectiveness of the CALNet of modularizing and assembling attributes, and showed the advantages of the CALNet in solving complicated tasks compared to the baseline RL. Our future work includes transferring attributes between different base attributes and even different agents. Another potential direction is to investigate the attribute learning models that can assemble lots of attributes. We believe that attribute learning can help human build versatile controllers more easily.
\begin{comment} 
Although cascade structure can ensure the attribute requirement and the base attribute, when the numbers of attribute increase, it becomes hard for attribute network to maintain the ability to accomplish the base task. So it is a very promising work to explore the structure for multi-attribute environment in the future.
\end{comment}

\begin{comment}
\section*{ACKNOWLEDGMENT}

The authors thank ...
\end{comment}

%%%%%%%%%%%%%%%%%%%%%%%%%%%%%%%%%%%%%%%%%%%%%%%%%%%%%%%%%%%%%%%%%%%%%%%%%%%%%%%%

\end{document}